\documentclass[
    journal=aas, % happens to use APA style, which is required
    manuscript=article-type,
    year=2026,
    volume=1
]{cup-journal}

\makeatletter
\patch@numeric@authors
\makeatother

\usepackage{xpatch}
\makeatletter
\xpatchcmd{\@maketitle}
  {\parbox[b]{\dimexpr\textwidth-26mm\relax}{%
    \firstheadsize{\journalnamefont\cup@journal@name} 
    (\cup@year), {\volumefont\cup@vol}, \thepage--\pageref{LastPage}
    \par% doi:{\cup@doi}
  }\hfill\par
  \vspace*{\baselineskip}}
  {}{}{}
\makeatother

\makeatletter
\xpatchcmd{\@maketitle}
  {{\sffamily\bfseries\color{structure@color}%
    \MakeUppercase{\convertchar{\cup@manuscript}{-}{ }}\par}}
  {}{}{}
\makeatother

\makeatletter
\renewcommand{\@oddhead}{\hfill\thepage}%
\makeatother

\colorlet{tbrowcolor}{white}
\colorlet{tbheadcolor}{white}

\usepackage{amsfonts}

\usepackage{graphicx}

\usepackage{booktabs}
\usepackage{ccicons}
\usepackage{fontawesome5} % 6 and 7 have thinner ?
\usepackage{makecell}
\usepackage{multirow}
\usepackage{tikz}

\newcommand{\negphantom}[1]{\settowidth{\dimen0}{#1}\hspace*{-\dimen0}}

\newcommand*\circled[1]{\tikz[baseline=(char.base)]{
            \node[shape=circle,draw,line width=0.1em,scale=0.45,inner sep=0.2em] (char) {#1};}}
\newcommand{\ccUnknown}{\circled{\faQuestion{}}}
\newcommand{\ccProprietary}{\circled{\resizebox{!}{0.85em}{\faLock{}}}}

\usepackage[frozencache]{minted}
\setminted{
  style=friendly,
  bgcolor=gray!10,
  baselinestretch=0.9,
  breaklines,
  breakanywhere,
  breakautoindent,
  breakindent=1em,
  xleftmargin=0.3em,
  tabsize=2,
  autogobble
}

\title{Earth Embeddings}

\author{Adam J. Stewart}
\affiliation{Chair of Data Science in Earth Observation, Technical University of Munich, Munich, Germany}
\alsoaffiliation{Munich Center for Machine Learning, Munich, Germany}
\email{adam.stewart@tum.de}

\author{Heng Fang}
\affiliation{KTH Royal Institute of Technology, Stockholm, Sweden}

\author{Isaac A. Corley}
\affiliation{Taylor Geospatial, San Antonio, USA}

\author{Xiao Xiang Zhu}
\affiliation{Chair of Data Science in Earth Observation, Technical University of Munich, Munich, Germany}
\alsoaffiliation{Munich Center for Machine Learning, Munich, Germany}

\begin{document}

\maketitle

\begin{abstract}
    Earth observation is moving from foundation models that users must run themselves toward embedding products that package model feature outputs as reusable data without needing to download and process the imagery used to generate them. Earth embeddings are vectors that summarize locations, image patches, or pixels, letting users analyze compact features instead of repeatedly training or running large models on raw satellite imagery. This chapter explains the main types of Earth embeddings, from implicit location encoders to explicit patch and pixel products, and compares their coverage, resolution, dimensionality, storage cost, licenses, and reproducibility. We review their use in land cover and crop mapping, ecological and hazard modeling, socioeconomic prediction, and semantic search, with evidence on when embeddings improve on conventional features and when pooling, fusion, or spatial transfer limit performance. Two case studies show practical workflows for similarity search and land cover mapping. We close with guidance for choosing, evaluating, storing, compressing, and publishing embeddings, and with open problems in oceanic and atmospheric coverage, uncertainty, and benchmarking.
\end{abstract}

\section{Introduction}

\emph{Foundation models} are commonly understood as models trained on broad data at scale and adaptable to many downstream tasks \citep{bommasani2021opportunities}. 
An \emph{embedding} is the vector representation produced by such a model or encoder: a compressed numerical description of an image, patch, pixel, location, or time period that can be reused for downstream analysis. In Earth observation (EO), this idea is attractive because public EO archives are already at petabyte scale and continue to grow rapidly: for example, NASA's EOSDIS reported 178.7 PB of archived Earth science data and an average archive growth of 160 TB/day for FY2025 \citep{nasa_esds_metrics_2025}. 
% Much of these data are richly georeferenced and calibrated, but not annotated with task-specific labels, and they are spatially structured, multispectral, and multitemporal. 
In this chapter, we use the term \textit{Earth embeddings} to refer to reusable vector representations of the Earth system, including implicit location embeddings, patch-level image embeddings, and dense pixel-level embedding products.

The idea of using embeddings as reusable representations predates EO. In natural language processing, Word2Vec showed that words could be represented as continuous vectors whose geometry captures semantic and syntactic relationships \citep{mikolov2013efficient}. Later, BERT demonstrated the value of contextual embeddings learned through self-supervised pretraining and adapted to many language tasks with limited task-specific architecture changes \citep{devlin2019bert}. In computer vision, CLIP extended this logic to multimodal alignment by learning a shared image--text embedding space that supports zero-shot transfer and retrieval \citep{radford2021learning}. These developments established the central promise of embeddings: expensive representation learning can be performed once, while downstream users operate on compact features rather than raw data.

EO adopted this idea more slowly because satellite data differ from natural images in important ways. Satellite observations are overhead, georeferenced, multispectral, multitemporal, and often multi-sensor; they also exhibit strong spatial autocorrelation, cloud contamination, irregular revisit patterns, and domain shifts across regions and seasons. \citet{rolf2024position} therefore argue that satellite data should be treated as a distinct machine-learning modality rather than as ordinary RGB imagery. Early EO representation learning work responded to this gap by developing in-domain self-supervised pretraining methods. Tile2Vec \citep{jean2019tile2vec} explored relative geographic distance between image triplets using contrastive margin loss, while GASSL \citep{ayush2021geography} explored geolocation as a pretext task. SeCo used seasonal contrastive learning over unlabeled remote-sensing data to improve transfer relative to ImageNet pretraining \citep{manas2021seasonal}, while SatMAE adapted masked autoencoders to temporal and multispectral satellite imagery \citep{cong2022satmae}. Prithvi later illustrated the foundation model framing more explicitly by pretraining a transformer on Harmonized Landsat Sentinel-2 imagery and adapting it to tasks such as flood mapping, wildfire scar segmentation, multitemporal crop segmentation, and multitemporal cloud-gap imputation \citep{jakubik2023foundation}. DOFA introduced the idea of using a wavelength-conditioned hypernetwork to dynamically generate the first layer of a model, allowing a single set of shared model weights to simultaneously support synthetic-aperture radar (SAR), red-green-blue (RGB), multispectral (MS), and hyperspectral (HS) imagery for the first time \citep{dofa}.

As EO foundation models proliferated, the community began to develop benchmarks for comparing them. GEO-Bench introduced six classification and six segmentation tasks for Earth monitoring and evaluated a range of baselines, including ResNet, ConvNeXt, ViT, SwinV2, SeCo, MoCo-S2, and DINO-S2 variants \citep{lacoste2023geo}. PANGAEA later broadened this effort with a more global and inclusive protocol covering diverse sensors, resolutions, temporal modalities, and tasks, and benchmarked or supported various geospatial foundation models such as SSL4EO-S12 variants, Scale-MAE, and DOFA against supervised baselines such as UNet and ViT \citep{marsocci2025pangaea}. Further, NeuCo-Bench introduced SSL4EO-S12-downstream, composed of 8 multispectral and multitemporal downstream tasks to evaluate the representation and compression ability of models pretrained on SSL4EO-S12 \citep{Vinge_2026_CVPR}. These benchmarks show both the promise and the limits of geospatial foundation models: pretraining can improve transfer, but current models do not consistently dominate supervised baselines across all tasks, sensors, and regions.

A newer development is the shift from foundation models as trainable backbones to \textit{embeddings as products}. MOSAIKS anticipated this direction by showing that a single satellite-image encoding could support diverse prediction tasks with much lower computational cost than task-specific deep learning \citep{mosaiks}. Recent concept papers make this shift explicit, arguing that precomputed embeddings can decouple downstream analysis from model-specific training and inference pipelines, but also create new challenges in standardization, access, resolution, and evaluation \citep{gomes2025lossy,fang2026earth,rolf2025earth,klemmer2025earth}. While foundation models are the source of many modern EO representations, this chapter focuses primarily on the embeddings they produce: reusable vector representations of locations, image patches, pixels, or annual surface conditions. This focus shifts attention from model architecture alone to the practical question of how representations are stored, accessed, evaluated, and reused in downstream geospatial analysis.

\section{Landscape}

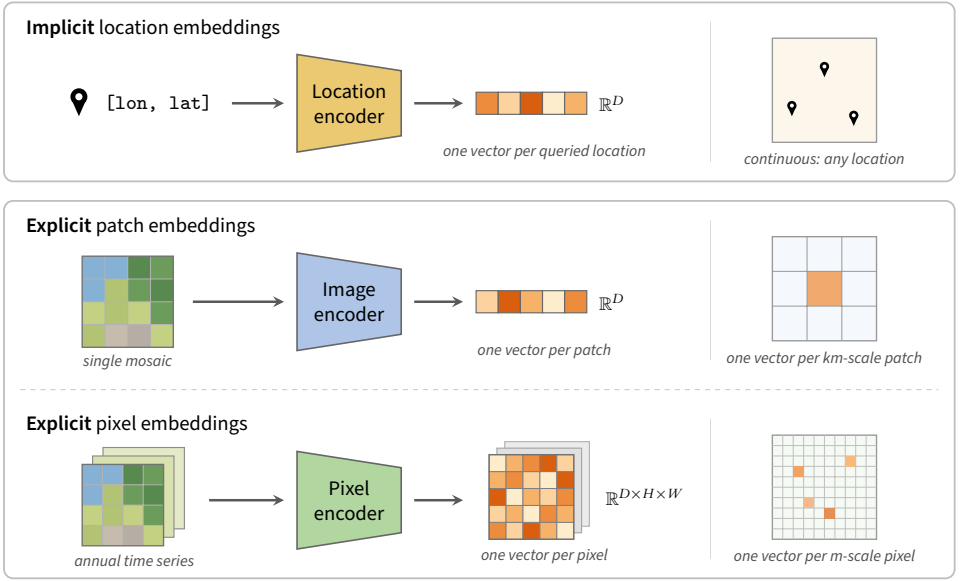
\begin{figure}[htbp]
    \centering
    % Earth embedding types: implicit location (top box), explicit patch + pixel (bottom box).
% Input with \input{figures/embedding-types} inside a figure environment;
% design width is 14.6cm, wrap in \resizebox{\textwidth}{!}{...}.
\resizebox{0.9\columnwidth}{!}{
\begin{tikzpicture}[
    x=1cm, y=1cm,
    font=\sffamily,
    every node/.style={inner sep=1pt},
    flow/.style={-stealth, line width=1.1pt, draw=black!65},
]
\definecolor{encgold}{RGB}{235,201,113}
\definecolor{encblue}{RGB}{174,197,232}
\definecolor{encgreen}{RGB}{178,214,160}
\definecolor{embA}{RGB}{217,95,14}
\definecolor{embB}{RGB}{236,140,60}
\definecolor{embC}{RGB}{247,178,106}
\definecolor{embD}{RGB}{252,211,158}
\definecolor{embE}{RGB}{254,237,205}
% land-cover palette for input mosaics (cell-aligned classes)
\definecolor{lcwater}{RGB}{132,177,212}
\definecolor{lcforestA}{RGB}{88,138,79}
\definecolor{lcforestB}{RGB}{108,156,92}
\definecolor{lcfield}{RGB}{196,209,130}
\definecolor{lcfieldB}{RGB}{178,196,116}
\definecolor{lcurban}{RGB}{198,188,174}
\definecolor{lcurbanB}{RGB}{182,174,164}

% ---- helpers -------------------------------------------------------------
% map pin at (#1,#2), scale #3
\newcommand{\pinat}[3]{
  \fill[black] ({#1-0.105*#3},{#2+0.24*#3}) -- ({#1+0.105*#3},{#2+0.24*#3}) -- (#1,#2) -- cycle;
  \fill[black] (#1,{#2+0.30*#3}) circle ({0.13*#3});
  \fill[white] (#1,{#2+0.30*#3}) circle ({0.055*#3});
}
% 1D embedding vector (5 cells), left edge at (#1,#2), cell colors #3
% (unordered, like the pixel grids: dimension values carry no ordering)
\newcommand{\embvec}[3]{
  \foreach \c [count=\i from 0] in {#3}{
    \draw[black!60, line width=0.5pt, fill=\c]
      ({#1+0.34*\i},{#2-0.17}) rectangle ({#1+0.34*\i+0.34},{#2+0.17});
  }
}
% encoder trapezoid, left edge at x=#1 centered on y=#2, fill #3, label #4
\newcommand{\encoder}[4]{
  \draw[black!60, line width=0.7pt, fill=#3]
    (#1,{#2-0.75}) -- ({#1+1.6},{#2-0.5}) -- ({#1+1.6},{#2+0.5}) -- (#1,{#2+0.75}) -- cycle;
  \node[align=center, font=\sffamily\small] at ({#1+0.8},{#2}) {#4};
}
% stylized satellite mosaic, lower-left (#1,#2), side #3
% 4x4 land-cover map, one class per grid cell: lake (top-left), forest block
% (right), fields (middle/left), built-up (bottom)
\newcommand{\mosaic}[3]{
  \foreach \c [count=\i from 0] in {lcwater,lcwater,lcforestA,lcforestB}
    \fill[\c] ({#1+0.25*#3*\i},{#2+0.75*#3}) rectangle ++({0.25*#3},{0.25*#3});
  \foreach \c [count=\i from 0] in {lcwater,lcfieldB,lcforestB,lcforestA}
    \fill[\c] ({#1+0.25*#3*\i},{#2+0.5*#3}) rectangle ++({0.25*#3},{0.25*#3});
  \foreach \c [count=\i from 0] in {lcfield,lcfieldB,lcfield,lcforestB}
    \fill[\c] ({#1+0.25*#3*\i},{#2+0.25*#3}) rectangle ++({0.25*#3},{0.25*#3});
  \foreach \c [count=\i from 0] in {lcfieldB,lcurban,lcurbanB,lcfield}
    \fill[\c] ({#1+0.25*#3*\i},{#2}) rectangle ++({0.25*#3},{0.25*#3});
  \draw[black!30, line width=0.3pt]
    ({#1+0.25*#3},#2) -- ++(0,#3) ({#1+0.5*#3},#2) -- ++(0,#3) ({#1+0.75*#3},#2) -- ++(0,#3)
    (#1,{#2+0.25*#3}) -- ++(#3,0) (#1,{#2+0.5*#3}) -- ++(#3,0) (#1,{#2+0.75*#3}) -- ++(#3,0);
  \draw[black!55, line width=0.6pt] (#1,#2) rectangle ++(#3,#3);
}
% panel box, top edge at y=#1, height #2, title #3
\newcommand{\panel}[3]{
  \draw[black!25, line width=0.8pt, rounded corners=4pt, fill=white]
    (0,#1) rectangle (14.6,{#1-#2});
  \node[anchor=west, font=\sffamily\small] at (0.3,{#1-0.4}) {#3};
}

% ==== row (a): implicit location embeddings ================================
\panel{0}{2.75}{\textbf{Implicit} location embeddings}
\pinat{1.15}{-1.75}{1.0}
\node[anchor=west, font=\ttfamily\small] at (1.5,-1.55) {[lon, lat]};
\draw[flow] (3.5,-1.55) -- (4.3,-1.55);
\encoder{4.5}{-1.55}{encgold}{Location\\encoder}
\draw[flow] (6.3,-1.55) -- (7.05,-1.55);
\embvec{7.25}{-1.55}{embB,embD,embA,embE,embC}
\node[anchor=west, font=\footnotesize] at (9.1,-1.55) {$\mathbb{R}^{D}$};
\node[font=\scriptsize\itshape, text=black!70] at (8.3,-2.3) {one vector per queried location};
\draw[black!15] (10.85,-0.3) -- (10.85,-2.45);
\draw[black!40, line width=0.6pt, fill=encgold!15] (11.8,-2.15) rectangle (13.4,-0.55);
\pinat{12.1}{-1.75}{0.55}
\pinat{12.6}{-1.15}{0.55}
\pinat{13.05}{-1.9}{0.55}
\node[font=\scriptsize\itshape, text=black!70] at (12.6,-2.42) {continuous: any location};

% ==== rows (b)+(c): explicit embeddings, one shared panel ==================
\panel{-3.05}{5.8}{\textbf{Explicit} patch embeddings}
\mosaic{1.2}{-5.3}{1.4}
\node[font=\scriptsize\itshape, text=black!70] at (1.9,-5.52) {single mosaic};
\draw[flow] (2.9,-4.6) -- (4.3,-4.6);
\encoder{4.5}{-4.6}{encblue}{Image\\encoder}
\draw[flow] (6.3,-4.6) -- (7.05,-4.6);
\embvec{7.25}{-4.6}{embD,embA,embC,embE,embB}
\node[anchor=west, font=\footnotesize] at (9.1,-4.6) {$\mathbb{R}^{D}$};
\node[font=\scriptsize\itshape, text=black!70] at (8.3,-5.35) {one vector per patch};
\draw[black!15] (10.85,-3.35) -- (10.85,-5.5);
\draw[black!40, line width=0.6pt, fill=encblue!12] (11.8,-5.2) rectangle (13.4,-3.6);
\fill[embB!75] (12.333,-4.667) rectangle (12.867,-4.133);
\draw[black!40, line width=0.35pt]
  (12.333,-5.2) -- (12.333,-3.6) (12.867,-5.2) -- (12.867,-3.6)
  (11.8,-4.667) -- (13.4,-4.667) (11.8,-4.133) -- (13.4,-4.133);
\node[font=\scriptsize\itshape, text=black!70] at (12.6,-5.47) {one vector per km-scale patch};

% ---- row (c): explicit pixel embeddings (inside shared panel) -------------
\draw[black!20, line width=0.5pt, dash pattern=on 2.2pt off 2.2pt]
  (0.25,-5.95) -- (14.35,-5.95);
\node[anchor=west, font=\sffamily\small] at (0.3,-6.5) {\textbf{Explicit} pixel embeddings};
% time-series stack (back to front)
\draw[black!45, line width=0.5pt, fill=lcfield!45] (1.52,-8.09) rectangle ++(1.25,1.25);
\draw[black!45, line width=0.5pt, fill=lcfield!65] (1.36,-8.22) rectangle ++(1.25,1.25);
\mosaic{1.2}{-8.35}{1.25}
\node[font=\scriptsize\itshape, text=black!70] at (2.0,-8.57) {annual time series};
\draw[flow] (3.15,-7.65) -- (4.3,-7.65);
\encoder{4.5}{-7.65}{encgreen}{Pixel\\encoder}
\draw[flow] (6.3,-7.65) -- (7.05,-7.65);
% dense embedding grid with depth
\draw[black!30, line width=0.5pt, fill=black!8] (7.69,-8.05) rectangle ++(1.3,1.3);
\draw[black!35, line width=0.5pt, fill=black!12] (7.57,-8.15) rectangle ++(1.3,1.3);
\foreach \c [count=\i from 0] in {embB,embD,embA,embC,embE}
  \draw[black!50, line width=0.3pt, fill=\c] ({7.45+0.26*\i},-8.25) rectangle ++(0.26,0.26);
\foreach \c [count=\i from 0] in {embD,embC,embE,embB,embC}
  \draw[black!50, line width=0.3pt, fill=\c] ({7.45+0.26*\i},-7.99) rectangle ++(0.26,0.26);
\foreach \c [count=\i from 0] in {embA,embE,embC,embD,embB}
  \draw[black!50, line width=0.3pt, fill=\c] ({7.45+0.26*\i},-7.73) rectangle ++(0.26,0.26);
\foreach \c [count=\i from 0] in {embC,embB,embD,embE,embA}
  \draw[black!50, line width=0.3pt, fill=\c] ({7.45+0.26*\i},-7.47) rectangle ++(0.26,0.26);
\foreach \c [count=\i from 0] in {embE,embC,embB,embA,embD}
  \draw[black!50, line width=0.3pt, fill=\c] ({7.45+0.26*\i},-7.21) rectangle ++(0.26,0.26);
\draw[black!55, line width=0.6pt] (7.45,-8.25) rectangle (8.75,-6.95);
\node[anchor=west, font=\footnotesize] at (9.2,-7.6) {$\mathbb{R}^{D \times H \times W}$};
\node[font=\scriptsize\itshape, text=black!70] at (8.3,-8.5) {one vector per pixel};
\draw[black!15] (10.85,-6.4) -- (10.85,-8.55);
\draw[black!40, line width=0.6pt, fill=encgreen!12] (11.8,-8.25) rectangle (13.4,-6.65);
\foreach \i in {1,...,9}
  \draw[black!25, line width=0.25pt]
    ({11.8+0.16*\i},-8.25) -- ({11.8+0.16*\i},-6.65)
    (11.8,{-8.25+0.16*\i}) -- (13.4,{-8.25+0.16*\i});
\fill[embB!80] (12.12,-7.29) rectangle ++(0.16,0.16);
\fill[embA!70] (12.6,-7.93) rectangle ++(0.16,0.16);
\fill[embC!90] (12.92,-7.13) rectangle ++(0.16,0.16);
\fill[embB!60] (12.28,-7.77) rectangle ++(0.16,0.16);
\node[font=\scriptsize\itshape, text=black!70] at (12.6,-8.52) {one vector per m-scale pixel};
\end{tikzpicture}
}
    \caption{\textbf{Three families of Earth embeddings.} Implicit location encoders (top) map coordinates directly to an embedding, requiring no imagery at inference time and supporting queries at any location. Explicit patch embeddings (middle) encode a single image mosaic into one vector per km-scale tile. Explicit pixel embeddings (bottom) encode an annual image time series into a dense field with one vector per pixel. Figure design adapted from \citet{klemmer2025earth}.}
    \label{fig:embedding-types}
\end{figure}

As of July 2026, there is a rich diversity of embedding techniques, models, and products. We distinguish between two broad families of embeddings: implicit and explicit (see Figure~\ref{fig:embedding-types}). \emph{Implicit} models are capable of generating an embedding representation solely from location information. These models are often jointly trained with \emph{explicit} models, which generate an embedding representation from input satellite image data. Explicit embeddings are further divided by the spatial unit of the output: patch embeddings summarize an entire image mosaic with a single vector, while pixel embeddings retain a vector for each input pixel.

\subsection{Implicit embeddings}

Implicit neural representations are a powerful strategy for encoding information directly into a model, removing the need for input imagery at inference time. They generally involve two separate models: an image encoder,
\begin{equation}
    f_\mathrm{img}: \mathbb{R}^{C \times H \times W} \rightarrow \mathbb{R}^D,
\end{equation}\label{eqn:image}% <- this comment is actually needed to suppress a space
and a location encoder,
\begin{equation}
    f_\mathrm{loc}: (x, y) \rightarrow \mathbb{R}^D.
\end{equation}\label{eqn:location}% <- this comment is actually needed to suppress a space
These two encoders are then jointly trained on geotagged imagery such that both models generate similar embedding representations for a given input. This training can be done either through knowledge distillation, where features from a pretrained teacher network (the image encoder) are distilled to a smaller student network (the location encoder), or through contrastive learning, where the models learn to generate similar representations for positive samples and different representations for negative samples. Here, positive samples denote images with matching locations, negative samples denote images with non-matching locations, and similarity is usually defined as cosine similarity in the embedding space. At runtime, the image encoder is typically discarded and the location encoder implicitly stores all learned representations.

\begin{table}[htbp]
    \centering
    \caption{\textbf{Implicit embeddings available as of July 2026.} Includes dimensionality, model architecture, and training technique. Implicit embeddings are typically generated by a location encoder jointly trained with an image encoder. Location encoders typically include a positional encoder such as Random Fourier Features (RFF) or Spherical Harmonics (SH).}
    \label{tab:location}
    \begin{tabular}{lrrcccc}
        \toprule
        \textbf{Name} & \textbf{Year} & \textbf{Dim.} & \textbf{Image Encoder} & \textbf{Location Encoder} & \textbf{Positional Encoder} & \textbf{Training} \\
        \midrule
        GPS2Vec & 2019 & 2000 & SIFT & MLP & Grid & Distillation \\
        GPS2Vec+ & 2021 & 1365 & VGG-16 & MLP & Grid & Distillation \\
        CSP & 2023 & 512 & \makecell{InceptionV3 \\ ResNet-50} & MLP & Grid & Contrastive \\
        SINR & 2023 & 256 & - & Residual MLP & Sinusoidal & Supervised \\
        GeoCLIP & 2023 & 512 & ViT-L/14 & MLP & EEP + RFF & Contrastive \\
        SatCLIP & 2023 & 256 & \makecell{ResNet-18 \\ ResNet-50 \\ ViT-S/16} & SIREN & SH & Contrastive \\
        TaxaBind & 2024 & 512 & ViT-B/16 & Residual MLP & EEP + RFF & Contrastive \\
        RANGE & 2025 & 1280 & ViT-S/16 & SIREN & SH & Contrastive \\
        GAIR & 2025 & 768 & ViT-B/16 & MLP & EEP + RFF & Contrastive \\
        Climplicit & 2025 & 256 & ResNet-18 & ReSIREN & Sinusoidal & Contrastive \\
        TIGeR & 2026 & 1024 & ViT-L/14 & MLP & HEALPix + RFF & Contrastive \\
        LIANet & 2026 & 4 & \makecell{ResNet-50 + U-Net \\ ResNet-101 + U-Net} & Hash Table & Grid & Generative \\
        UniGeoCLIP & 2026 & 768 & ViT-B/16 & Transformer & EEP + RFF & Contrastive \\
        TTE & 2026 & 512 & ViT-L/16 & MLP & Spherical Voronoi & Contrastive \\
        \bottomrule
    \end{tabular}
\end{table}

The evolution of image encoders in implicit embedding generation has largely followed that of the broader computer vision community. Early works like GPS2Vec \citep{yin2019gps2vec} relied on bag-of-visual-words feature extractors like Scale-Invariant Feature Transform (SIFT) \citep{lowe1999object}. GPS2Vec+ \citep{yin2021learning} replaced this SIFT extractor with a VGG-16 network \citep{simonyan2014very}. Contrastive Spatial Pretraining (CSP) \citep{mai2023csp} explored both InceptionV3 \citep{szegedy2015going} and ResNet-50 \citep{he2016deep} architectures. Spatial Implicit Neural Representations (SINR) lacked an image encoder entirely, instead opting for supervised learning on species occurrence data \citep{cole2023spatial}. All later works use either a ResNet or Vision Transformer (ViT) architecture \citep{dosovitskiy2020image}.

The evolution of location encoders is far more interesting and unique. Early work on location encoders found that raw \((x, y)\) locations resulted in overly smooth embeddings, edge artifacts, and latitudinal bias, necessitating the inclusion of positional encoders to refine the input. Early works like GPS2Vec and CSP relied on a simple Multi-Layer Perceptron (MLP) \citep{rumelhart1986learning}, learning a separate model for each UTM zone to increase spatial granularity. SINR introduced the use of a Residual MLP and sinusoidal positional encoder to improve gradient flow and avoid edge artifacts, respectively. After that, two parallel works inspired by Contrastive Language--Image Pre-training (CLIP) \citep{radford2021learning} emerged.

GeoCLIP \citep{vivanco2023geoclip} combined an MLP with an Equal Earth Projection (EEP) \citep{vsavrivc2019equal} and Random Fourier Features (RFF) \citep{tancik2020fourier} to alleviate latitudinal and spectral biases, respectively. This same strategy was later adopted by TaxaBind \citep{sastry2025taxabind} and Geo-Aligned Implicit Representations (GAIR) \citep{liu2026gair}, alongside additional encoders for other modalities. TIGeR \citep{shatwell2026tiger} later replaced EEP with a HEALPix projection \citep{gorski2005healpix}, and UniGeoCLIP \citep{astruc2026unigeoclip} replaced the MLP with a Transformer \citep{vaswani2017attention}.

SatCLIP \citep{klemmer2025satclip} took a different direction, using a Sinusoidal Representation Network (SIREN) architecture \citep{sitzmann2020implicit}, which uses Spherical Harmonics (SH) to learn representations at different scales. RANGE later adopted this same architecture, showing that Retrieval-Augmented Generation (RAG) can improve the spatial resolution of implicit neural representations \citep{dhakal2025range}. Climplicit added residual connections to SIREN, creating a novel ReSIREN architecture \citep{dollinger2025climplicit}.

Location Is All You Need Network (LIANet) represents a newer direction, replacing the image \emph{encoder} with an image \emph{decoder} and generating output imagery from only a learned low-dimensional gridded hash table lookup \citep{lianet}. Tessellating the Earth (TTE) explores spherical Voronoi cells for position encoding, allowing for dynamic tessellation of diverse landscapes and larger shared cells for uniform ocean surfaces \citep{cher2026tessellating}. Table~\ref{tab:location} gives a rundown of several prominent implicit embeddings, their embedding dimensions, model architectures, and training techniques.

\subsection{Explicit embeddings}

While location encoders allow global information to be compressed within a small model architecture, they typically lack the fine-grained spatial resolution needed for many tasks in remote sensing. Many explicit embeddings are created using an image encoder instead of simply discarding it. Unlike location encoders, inference of pretrained models is run once globally, producing a static embedding product that can be versioned and reused by others without the technical expertise or compute requirements of deep learning. We further subdivide these explicit embeddings based on their input/output dimensions, spatial resolution, and target applications.

\begin{table}[htbp]
    \centering
    \caption{\textbf{Patch embedding products available as of July 2026.} Temporal resolution is divided into ``snapshot'' for embeddings generated from a single mosaic and ``annual'' for embeddings generated from annual time series data. *Product has sparse spatial or temporal coverage.}
    \begin{tabular}{lllrlcrr}
        \toprule
        & & \multicolumn{2}{c}{\textbf{Spatial}} & \multicolumn{2}{c}{\textbf{Temporal}} \\
        \cmidrule(l){3-4} \cmidrule(l){5-6}
        \textbf{Family} & \textbf{Subproduct} & \textbf{Extent} & \textbf{Resolution} & \makecell[c]{\textbf{Extent}} & \textbf{Resolution} & \textbf{Dim.} & \textbf{Dtype} \\
        \midrule
         & USA grid & USA & 1~km & \phantom{2018--}2018 & & 8192 & \\
        \multirow{-2}{*}{\makecell{MOSAIKS \\ (2021, 2025)}} & Global grid & Global & 0.01\textdegree\negphantom{\textdegree}\phantom{ km} & \phantom{2019--}2019 & \multirow{-2}{*}{Snapshot} & 4000 & \multirow{-2}{*}{float32} \\
        \midrule
         & v0 Sentinel & Global* & 5.12~km & 2018--2023* & & 768 & \\
         & v1.5 NAIP & USA & 154--256~m\phantom{k} & 2010--2021* & Snapshot & 1024 & float32 \\
        \multirow{-3}{*}{\makecell{Clay \\ (2024, 2026)}} & v1.5 LGND & Global & 2.56~km & 2024--2025 & & 1024 & \\
        \midrule
         & Quasara-DE & Germany & & & & & \\
         & Quasara-EU & Europe & \multirow{-2}{*}{3.84~km} & \multirow{-2}{*}{2020--2024*} & \multirow{-2}{*}{Snapshot} & \multirow{-2}{*}{1152} & \multirow{-2}{*}{float32} \\
         \cmidrule(l){2-8}
         & Core-SSL4EO-S2 & & 2.24~km & & & 2048 \\
         & Core-SSL4EO-S1 & & 2.24~km & & & 2048 \\
         & Core-DINOv2 & & 2.24~km & & & 2048 \\
         & Core-SigLIP & & 3.84~km & & & 2048 \\
         & Core-DeCUR-S2 & & 2.24~km & & & 2048 \\ 
         & Core-DeCUR-S1 & & 2.24~km & & & 2048 \\
         & Core-MMEarth & & 80--800~m\phantom{k} & & & 320 \\
         & Core-UniverSat & \multirow{-8}{*}{Global} & 120~m\phantom{k} & \multirow{-8}{*}{2016--2024*} & \multirow{-8}{*}{Snapshot} & 768 & \multirow{-8}{*}{float32} \\ 
         \cmidrule(l){2-8}
         & 249k-Clay & & & & & 1024 \\
         & 249k-OlmoEarth & & & & & 768 \\
         & 249k-SatCLIP & & & & & 256 \\
         & 249k-DINOv2 & & & & & 1024 \\
         & 249k-FarSLIP & & & & & 512 \\
        \multirow{-16}{*}{\makecell{Major TOM \\ (2024--2026)}} & 249k-SigLIP & \multirow{-6}{*}{Global*} & \multirow{-6}{*}{3.84~km} & \multirow{-6}{*}{2016--2024*} & \multirow{-6}{*}{Snapshot} & 1152 & \multirow{-6}{*}{float32} \\
        \midrule
        \multicolumn{2}{l}{Earth Index Embeddings (2025)} & Global & 320~m\phantom{k} & \phantom{2024--}2024 & Snapshot & 384 & float32 \\
        \midrule
        \multicolumn{2}{l}{Copernicus-Embed (2025)} & Global & 0.25\textdegree\negphantom{\textdegree}\phantom{ km} & \phantom{2021--}2021 & Annual & 768 & float32 \\
        \bottomrule
    \end{tabular}
    \label{tab:patch}
\end{table}

Patch embeddings typically involve a static mosaic image snapshot as input and a single 1D embedding as output to represent a large spatial region. This matches our mathematical definition in Equation~\ref{eqn:image}. These output embeddings are typically global and have a spatial resolution measured in kilometers or degrees. These medium-resolution embeddings can be used for land cover mapping, but are more commonly used for tasks like search and retrieval. Table~\ref{tab:patch} gives a rundown of all known patch embedding products as of July 2026.

The earliest patch embedding products were created using MOSAIKS, an unsupervised model based on Random Convolutional Features (RCF) \citep{mosaiks}. Clay also released embeddings from Clay v0 on its sparse global pretraining dataset. Later, it released higher-resolution embeddings from Clay v1.5 on NAIP data across the United States and Sentinel data with dense global coverage \citep{clay}.

The most influential patch embedding effort came from ESA \(\Phi\)-lab's Major TOM project. Major TOM started as a composable and expandable dataset, with a robust specification on how to organize and sample data to ensure gridded global coverage \citep{majortom}. Early community projects used Major TOM to build embeddings for Germany and Europe, while the Major TOM authors later released their own embeddings of the full global dataset \citep{majortomembeddings}. Small subsets of these global datasets with only 249k images were later used to generate embeddings for the EarthEmbeddingExplorer web application \citep{zheng2026earthembeddingexplorer}. 

Earth Index Embeddings were created with a similar intent, supporting relatively high-resolution search and retrieval with global coverage \citep{earthindex}. Copernicus-Embed represents a unique and interesting direction, generating embeddings from satellite images of the land surface, shallow ocean, and atmosphere \citep{copernicus}. The relatively low 0.25\textdegree resolution is designed for coupling with ERA5 data to support the use of high-resolution elevation and surface models in weather forecasting \citep{era5}.

\begin{table}[htbp]
    \centering
    \caption{\textbf{Pixel embedding products available as of July 2026.} Temporal resolution is divided into ``snapshot'' for embeddings generated from a single mosaic and ``annual'' for embeddings generated from annual time series data. *Product has sparse spatial or temporal coverage.}
    \begin{tabular}{@{}llrlcrr@{}}
        \toprule
        & \multicolumn{2}{c}{\textbf{Spatial}} & \multicolumn{2}{c}{\textbf{Temporal}} \\
        \cmidrule(l){2-3} \cmidrule(l){4-5}
        \textbf{Product} & \textbf{Extent} & \textbf{Resolution} & \makecell[c]{\textbf{Extent}} & \textbf{Resolution} & \textbf{Dim.} & \textbf{Dtype} \\
        \midrule
        Presto Embeddings (2025) & Togo & 10~m & 2019--2020 & Annual & 128 & uint16 \\
        Tessera Embeddings (2025) & Global* & 10~m & 2017--2025* & Annual & 128 & int8 \(\rightarrow\) float32 \\
        Google Satellite Embedding (2025) & Global & 10~m & 2017--2025 & Annual & 64 & int8 \(\rightarrow\) float64 \\
        Embedded Seamless Data (2026) & Global & 30~m & 2000--2024 & Annual & 12 & uint16 \(\rightarrow\) float32 \\
        \bottomrule
    \end{tabular}
    \label{tab:pixel}
\end{table}

In contrast to patch embeddings, pixel embeddings typically involve annual time series satellite imagery as input and high-resolution 1D embeddings as output for each input pixel. Their higher spatial resolution---and therefore storage cost---has necessitated clever ways to reduce dimensionality and compress data types during storage. Table~\ref{tab:pixel} gives a rundown of all known pixel embedding products as of July 2026.

Two families of pixel embedding model architectures have evolved. 1D pixel models focus on time series modeling while neglecting spatial correlations. These models typically follow the form:
\begin{equation}
    f_{\mathrm{1D}}: \mathbb{R}^{T \times C} \rightarrow \mathbb{R}^D.
\end{equation}
Presto \citep{presto} was one of the earliest 1D time series architectures used to generate pixelwise embeddings for the country of Togo \citep{prestoembed}. Tessera introduced one of the first global 1D pixelwise embedding products, with global coverage from Tessera v1.0 for 2024 and sparse on-demand coverage for Tessera v1.1 and other years \citep{tessera}.

3D pixel models incorporate both spatial and temporal inputs, generating dense pixelwise embeddings with spatiotemporal context. These models typically follow the form:
\begin{equation}
        f_{\mathrm{3D}}: \mathbb{R}^{T \times C \times H \times W} \rightarrow \mathbb{R}^{D \times H \times W}.
\end{equation}
Google released an early dense satellite embeddings product (GSE) using a proprietary AlphaEarth Foundations (AEF) model \citep{alphaearth} that fully incorporates both spatial and temporal information, with dense global coverage across the full history of Sentinel missions. Most recently, Embedded Seamless Data (ESD) \citep{esd} released embeddings using a proprietary ESDNet model, with close to 25 years of dense global coverage thanks to the long history of the Landsat missions.

\section{Current Usage}

% Geospatial embeddings are currently being adopted primarily as reusable representations rather than as stand-alone prediction systems. 
In practice, geospatial embeddings usually serve as input features for downstream models or as indexes for search and retrieval, replacing hand-crafted covariates, raw imagery, or task-specific preprocessing pipelines. Their use is most visible in applications where labeled data are sparse, input data are heterogeneous, or large-scale preprocessing of remote-sensing time series would otherwise be prohibitively expensive.

The clearest evidence so far comes from land cover, agriculture, and forest mapping. \citet{prestoembed} use Presto and GSE with a Random Forest classifier \citep{breiman2001random} for cropland mapping in Togo, where Presto achieves the best F1 score among the tested products. \citet{ishikawa2025assessing} use Presto, GSE, and Tessera embeddings for tree species classification in the Dutch National Forest Inventory \citep{francini2024forest}, reporting gains of 2--9\% points over hand-designed satellite time-series features in a sparse-label setting. \citet{ma2026harvesting} benchmark GSE for crop yield prediction, tillage mapping, and cover crop mapping in the United States, finding that the embeddings are competitive when models are trained and evaluated locally, but less robust under spatial transfer. 

A second set of applications uses embeddings as environmental covariates for physical and ecological mapping. Here, embeddings are used to summarize environmental conditions relevant to hazards, terrain, habitats, or ecosystem change, rather than directly mapping land cover or crops. \citet{cheng2026landslide} use GSE for landslide susceptibility mapping across Taiwan, Hong Kong, and Italy, showing that the full 64-dimensional representation outperforms conventional landslide conditioning factors in both F1 score and AUC. Similarly, \citet{hamoudzadeh2026inferring} show that GSE contains information about surface height, although prediction errors reveal remaining regional bias and distribution shift. In ecological applications, \citet{heiman2026characterizing} use GSE trajectories to characterize restoration outcomes in the Brazilian Atlantic Forest, while \citet{joseph2026continuous} use Clay embeddings to construct continuous biome representations that improve species occurrence modeling relative to discrete biome labels. Together, these studies suggest that embeddings can encode environmental structure beyond standard categorical maps or manually selected covariates, although the strength of this evidence varies by task and evaluation design.

Beyond biophysical mapping, embeddings are increasingly being used as general representations of places. In socioeconomic applications, \citet{pettersson2025leveraging} use GSE with graph neural networks for poverty mapping in Sub-Saharan Africa, showing that embeddings can substantially reduce the storage and preprocessing burden associated with raw Sentinel-2 imagery while still supporting predictive modeling. A different but related use case is search and retrieval, where embeddings are used to organize Earth observation data by semantic similarity rather than by predefined map classes. EarthEmbeddingExplorer demonstrates this direction using the Major TOM embeddings dataset to support interactive retrieval from text, image, and location queries \citep{zheng2026earthembeddingexplorer}. Similarly, work on GSE geometry shows that retrieval over local neighborhoods in embedding space can support environmental reasoning, while simple vector arithmetic is less reliable because the embedding manifold is not globally linear \citep{rahman2026characterizing}. These examples show that geospatial embeddings are not limited to supervised classification: they can also function as compact place descriptors and searchable indexes of the Earth's surface.

% The important point is that geospatial embeddings often do not replace Random Forests or other established models; they replace the input features. Many successful studies still train Random Forests, gradient-boosted trees, MLPs, graph neural networks, or U-Nets on top of embeddings. 
Earth embeddings do not remove the need for downstream modeling; instead, they move much of the deep-learning burden upstream into pretraining or embedding generation. As a result, many applications train comparatively lightweight models, including Random Forests, gradient-boosted trees, MLPs, graph neural networks, or U-Nets, on top of reusable embedding features. The Togo cropland study, for example, uses Presto embeddings as features in a Random Forest classifier \citep{prestoembed}, while the Dutch forest inventory study evaluates embeddings together with conventional classifiers and fine-tuning strategies \citep{ishikawa2025assessing}. This is an important practical advantage. Rather than training a foundation model or designing task-specific spectral--temporal features from scratch, users can often treat embeddings as analysis-ready covariates. This lowers the barrier for researchers who are familiar with GIS, geostatistics, remote sensing, or applied environmental modeling, but less familiar with large-scale model training and inference. The emergence of tools such as rs-embed further reflects this shift, aiming to make embeddings retrievable by region, time, and model through a common interface rather than through separate model-specific pipelines \citep{ye2026any}. 

At the same time, the evidence cautions against treating embeddings as universally superior representations. \citet{ma2026harvesting} find that GSE is competitive for several agricultural tasks when models are trained and evaluated locally, but is less reliable under spatial transfer and limited in its sensitivity to temporally specific agricultural signals. These results suggest that geospatial embeddings are working, but in a narrower and more practical sense than some foundation-model narratives imply. They are useful as compact, reusable feature layers and searchable representations, especially when labels are sparse or preprocessing is costly. They do not eliminate the need for spatial validation, temporal information, task-specific modeling, or classical learners. Their value depends on the embedding product, downstream task, transfer setting, and evaluation design.

\section{Case Studies}

Readers may wonder how to take advantage of embedding datasets to solve their own tasks in Earth observation and remote sensing. In this section, we walk through two common embedding applications: search and retrieval and land cover mapping. We show how to use existing embedding products and the model architectures used to create them with working code. These are expanded versions of case studies that first appeared in \citet{fang2026earth}.

All case studies are implemented using the TorchGeo library~\citep{torchgeo}. TorchGeo provides data loaders for all known embedding products. It also contains most of the components used to generate these embeddings:
\begin{itemize}
    \item \textbf{pretraining datasets}---Copernicus-Pretrain \citep{copernicus}, MMEarth \citep{mmearth}, SSL4EO \citep{ssl4eos12,ssl4eol},
    \item \textbf{SSL techniques}---BYOL \citep{byol}, MAE \citep{mae}, MoCo \citep{moco}, SimCLR \citep{simclr}, and
    \item \textbf{foundation models}---Copernicus-FM \citep{copernicus}, Presto \citep{presto}, Tessera \citep{tessera}
\end{itemize}

\subsection{Search and retrieval}

\begin{listing}[htbp]
\begin{minted}{python}
from torch.nn import CosineSimilarity
from torchgeo.datasets import EarthIndexEmbeddings, Sentinel2
from torchgeo.models import ViTSmall14_DINOv2_Weights, vit_small_patch14_dinov2

# User inputs
xmin, xmax, ymin, ymax = ...
xrad = (xmax - xmin) / 2
yrad = (ymax - ymin) / 2

# Datasets
earthindex = EarthIndexEmbeddings('data/earthindex')
sentinel2 = Sentinel2('data/sentinel2')
sample = sentinel2[xmin:xmax, ymin:ymax]
image = sample['image'] / 10_000  # normalization

# Models
model = vit_small_patch14_dinov2(ViTSmall14_DINOv2_Weights.SENTINEL2_ALL_SOFTCON)
embed = model(image)
cos = CosineSimilarity(dim=0)

# Search for most similar image patch
similarity = -1
x = y = 0
for sample in iter(earthindex):
    new_similarity = cos(embed, sample['embedding'])
    if new_similarity > similarity:
        similarity = new_similarity
        x, y = sample['x'], sample['y']

# Visualize resulting match
sample = sentinel2[x-xrad:x+xrad, y-yrad:y+yrad]
sentinel2.plot(sample)
\end{minted}
\caption{Example of a search and retrieval task in TorchGeo.}
\label{lst:search}
\end{listing}

Listing~\ref{lst:search} demonstrates an example of a search and retrieval embedding task implemented using TorchGeo. In this task, the goal is to look for regions around the world with a similar embedding representation to the reference region.

First, we select a reference region of interest in latitude/longitude coordinates. This also controls the size of each input patch. Next, we initialize a dataset for the Earth Index Embeddings and for one or more Sentinel-2 tiles. These datasets can live on the filesystem in a \texttt{data} directory or can be streamed directly from the cloud. We slice into the Sentinel-2 dataset to retrieve a GPU-ready PyTorch Tensor from our region of interest. We also divide the image by 10,000 to convert from digital numbers to reflectance values in a \([0, 1]\) range. We then initialize our model. We use the same ViT-S/14 DINOv2 model architecture and SoftCon weights \citep{softcon} used to generate the Earth Index Embeddings. We can easily use the same model to generate a new embedding for our reference image.

In order to find the most similar embedding, we need a definition of similarity. Here, we will use cosine similarity in embedding space as our distance metric. We can then iterate over all Earth Index Embeddings and search for one or more locations with similar embeddings. Note that while this simple example will work, we may want to use a \texttt{DataLoader} to construct mini-batches and move our objects to the GPU to speed up the similarity search. Once we have the most similar location, we can then feed this back into the original Sentinel-2 dataset, allowing us to quickly visualize the location in true-color RGB. This script may appear simple, but it could easily serve as the backend for a GUI such as a website.

\subsection{Land cover mapping}

\begin{listing}[htbp]
\begin{minted}{python}
from shapely import box
from torch.utils.data import DataLoader
from torchgeo.datasets import EuroCrops, TesseraEmbeddings, roi_split
from torchgeo.samplers import GriddedPatchSampler, RandomPatchSampler

# Datasets
tessera = TesseraEmbeddings('data/tessera')
eurocrops = EuroCrops('data/eurocrops', download=True)
dataset = tessera & eurocrops  # spatiotemporal intersection

# Geographic split
train_roi = box(-10, 35, 10, 60)  # Western Europe
test_roi  = box( 10, 35, 30, 60)  # Eastern Europe
train_dataset, test_dataset = roi_split(dataset, [train_roi, test_roi])

# Samplers
train_sampler = RandomPatchSampler(train_dataset, size=256)
test_sampler = GriddedPatchSampler( test_dataset, size=256, stride=128)

# Data loaders
train_dataloader = DataLoader(train_dataset, batch_size=256, sampler=train_sampler)
test_dataloader  = DataLoader( test_dataset, batch_size=256, sampler= test_sampler)

# Training
for epoch in range(10):
    for batch in train_dataloader:
        # Train a k-NN classifier or linear probe

# Inference
for batch in test_dataloader:
    # Make predictions and evaluate accuracy
\end{minted}
\caption{Example of a land cover mapping task in TorchGeo.}
\label{lst:map}
\end{listing}

Listing~\ref{lst:map} demonstrates an example of a land cover mapping task implemented using TorchGeo. In this task, the goal is to create a pixel-level map of land cover and/or land usage for a region of interest (ROI). More specifically, we will focus on crop type mapping across all of Europe.

First, we initialize a dataset for our Tessera embeddings. Note that we could also directly use Sentinel-2 imagery here, but it would require a larger deep learning model and more compute resources. We also initialize a dataset for our EuroCrops labels. EuroCrops is a vector dataset consisting of field boundaries and crop types for much of Europe \citep{eurocrops}. This dataset can be automatically downloaded and will automatically rasterize all vector shapefiles for us. Finally, we take the spatiotemporal intersection of these two datasets, as we only want matching regions and timespans with both embeddings and ground truth labels. We then perform a geographic split on our dataset. In this case, we will fine-tune our model on Western Europe and evaluate model performance on Eastern Europe. \texttt{roi\_split} ensures that there is no overlap between these train and test splits. Although we are using rectangular bounding boxes here for simplicity, any arbitrary Shapely \texttt{Polygon} will work.

Next, we initialize samplers that define the sampling strategy during training and testing. At training time, we choose random sampling in order to maximize data diversity. We will randomly sample \(256 \times 256\) pixel patches and construct mini-batches of these input and output pairs. At inference time, we do not want to randomly sample data. Instead, we want to ensure complete coverage of the entire region so we can create a map and evaluate its accuracy. To do this, we use a gridded sampling strategy, with a stride smaller than each patch size. This allows us to stitch together predictions and remove edge artifacts.

All datasets and samplers are wrapped by a \texttt{DataLoader}. We can then iterate over each data loader during training and testing. We can fit a scikit-learn classifier like \(k\)-NN or Random Forest or a PyTorch Linear layer to map from the embedding space to the smaller number of expected crop types. Once the shallow model is fine-tuned, we can make predictions and evaluate performance over all of Eastern Europe. Note that this kind of geographic split is common in geospatial machine learning and is required to evaluate out-of-domain performance.

\section{Experiments}

Direct comparison of geospatial embeddings is challenging due to differences in spatial unit, temporal support, and input modality among available products. Implicit embeddings encode locations, patch embeddings summarize image chips or mosaics, and pixel embeddings provide dense gridded representations. These differences make some embeddings better suited to spatial context, others to scene-level retrieval, and others to pixel- or region-level mapping. Consequently, there is no single best embedding across all downstream tasks; performance depends on spatial scale, label type, aggregation strategy, and temporal requirements.

This task dependence is already visible in one of the best-studied settings: scene classification. For these tasks, the strongest experimental evidence currently comes from explicit pixel embeddings. \citet{corley2026pixels} benchmark Google Satellite Embedding, Tessera, and OlmoEarth on EuroSAT-Embed and show that aggregation from pixels to patches is itself a major design choice: richer pooling strategies reduce the geographic generalization gap by more than 50\% relative to mean pooling and improve spatial-split accuracy by up to 6\%. This suggests that pixel embeddings are not automatically task-ready; their performance depends on how within-patch heterogeneity is summarized. Complementarity is also important. \Citet{van2026better} evaluate GSE, Tessera, GeoCLIP, and SatCLIP across six downstream tasks, finding that fused embeddings outperform the best single embedding in four of six cases. Thus, GSE and Tessera appear strong for land-cover-related information, while implicit embeddings such as GeoCLIP and SatCLIP can add complementary spatial context.

Accuracy alone also hides important task-specific structure within embeddings. For land-cover classification, \citet{benavides2026earth} analyze the public GSE product and characterize how individual embedding dimensions contribute to land-cover classes. Using feature importance patterns and progressive ablation, they find that 98\% of baseline land-cover classification performance can often be recovered with only 2 to 12 of the 64 dimensions, depending on the class. This suggests that, for land-cover classification specifically, the embedding space is functionally organized but also substantially redundant. Related diagnostic work by \citet{rao2025measuring} shows that geographic implicit neural representations can have intrinsic dimensions far below their ambient embedding dimensions, and that intrinsic dimensionality can correlate with downstream performance and reveal spatial artifacts. These results suggest that embedding evaluation should consider not only predictive accuracy but also task-specific redundancy, interpretability, and spatial structure.

Beyond supervised mapping, embeddings may also support similarity search and retrieval, but this use case is less systematically evaluated. \citet{betti2026s} show that implicit location embeddings, including GeoCLIP, SatCLIP, Climplicit, and CSP-fMoW, can be decomposed into human-interpretable representations while retaining high reconstruction capability, revealing interpretable geographic structures such as forests, deserts, roads, landmarks, and urban features. This suggests that implicit embeddings contain monosemantic geographic structure that could support retrieval and geographic search. However, interpretability does not by itself establish retrieval accuracy, and the current evaluation literature still lacks a common benchmark comparing patch, pixel, and implicit embeddings under shared retrieval metrics. Current evidence is therefore stronger for supervised classification and regression than for search.

A separate question is whether embeddings preserve temporal information needed for downstream use, rather than only spatial or semantic similarity. Here the evidence is weaker. Many products are annual or snapshot embeddings, which may be useful for mapping annual patterns such as phenology; however, most benchmarks collapse them into static representations before prediction. This matters because annual summaries can obscure both event timing in change detection and hemispheric differences in agricultural cycles. For example, an embedding aligned to one seasonal calendar may be much less informative for crop mapping in regions with different planting and harvest windows. \citet{gong2026earth} evaluate GSE and Clay for urban indicators from 2020 to 2023 and find that performance is more stable across years than across cities, suggesting some temporal robustness but not true change detection capabilities. Existing benchmarks provide little direct evidence for flooding, deforestation, event-level change detection, or temporally sensitive agricultural monitoring. The main conclusion is therefore cautious: current embeddings are experimentally strongest for land-cover-like mapping and spatial prediction, promising but under-benchmarked for semantic search, and still weakly evaluated for time-series tasks where the timing and direction of change are central.

\section{Reproducibility}

Scientific progress relies on the reproducibility of experiments. If the full embedding pipeline---including pretraining, inference, and embedding data and model code and weights---is unavailable or locked behind restrictive licenses, scientists risk dependence on opaque and uncertain company products. Furthermore, scientific progress can be gatekept by those with resources, making it harder for researchers to explore novel ideas and modifications of foundation models without recreating the entire pipeline from scratch. Reproducibility should be not only technically possible (open data, open code, open weights) but also as easy as possible \citep{heil2021reproducibility}.

To this end, we present a deep dive into the full pretraining and inference pipeline behind the most popular embeddings. We organize product licenses as follows:
\begin{itemize}
    \item \ccPublicDomain{} \textbf{public domain}: no copyright restrictions
    \item \ccAttribution{} \textbf{attribution}: must cite source
    \item \ccShareAlike{} \textbf{share-alike}: derived products must be shared under the same license
    \item \ccNoDerivatives{} \textbf{no derivatives}: derived products are not permitted
    \item \ccNonCommercial{} \textbf{non-commercial}: commercial usage is not permitted
    \item \ccUnknown{} \textbf{unknown}: source is available but no permissions are granted
    \item \ccProprietary{} \textbf{proprietary}: source is not publicly available
\end{itemize}
Public domain licenses like CC0-1.0 generally have the fewest restrictions. Permissive licenses like MIT and CC-BY-4.0 only require attribution. Copyleft licenses like GPL and CC-BY-SA-4.0 require derived products to be distributed under the same license. All other licenses place restrictions on how the product is used (CC-BY-ND-4.0) or who is allowed to use it (CC-BY-NC-4.0) and are considered ``source available'' but not ``open source''. Proprietary code or data is neither source available nor open source and prohibits full reproducibility.

\subsection{Location encoders}

\begin{table}[htbp]
    \centering
    \caption{\textbf{Reproducibility of location encoders as of July 2026.} Licenses are broken down into the license for the model code, weights, and data used to pretrain the model.}
    \begin{tabular}{lcccccc}
        \toprule
         & & & & \multicolumn{3}{c}{\textbf{Licenses}} \\
        \cmidrule(l){5-7}
        \textbf{Name} & \textbf{Model} & \textbf{Training} & \textbf{Data} & \textbf{Code} & \textbf{Weights} & \textbf{Data} \\
        \midrule
        GPS2Vec & SIFT + MLP & Distillation & 1M Flickr & \ccUnknown{} & \ccUnknown{} & \ccUnknown{} \\
        GPS2Vec+ & VGG-16 + MLP & Distillation & YLI-GEO & \ccUnknown{} & \ccUnknown{} & \ccPublicDomain{} \\
        CSP & \makecell{InceptionV3 + MLP \\ ResNet-50 + MLP} & Contrastive & \makecell{iNat2018 \\ fMoW} & \ccUnknown{} & \ccUnknown{} & \makecell{\ccAttribution{}\ccShareAlike{}\ccNoDerivatives{}\ccNonCommercial{} \\ \ccAttribution{}\ccShareAlike{}} \\
        SINR & Residual MLP & Supervised & iNat Open Data & \ccAttribution{} & \ccUnknown{} & \ccUnknown{} \\
        GeoCLIP & ViT-L/14 + MLP & Contrastive & MP-16 & \ccAttribution{} & \ccAttribution{} & \ccAttribution{}\ccShareAlike{}\ccNoDerivatives{}\ccNonCommercial{} \\
        SatCLIP & \makecell{ResNet-18 + SIREN \\ ResNet-50 + SIREN \\ ViT-S/16 + SIREN} & Contrastive & S2-100k & \ccAttribution{} & \ccAttribution{} & \ccAttribution{} \\
        TaxaBind & ViT-B/16 + MLP & Contrastive & iSatNat & \ccAttribution{} & \ccAttribution{} & \ccAttribution{}\ccNonCommercial{}\ccShareAlike{} \\
        RANGE & ViT-S/16 + SIREN & Contrastive & S2-100k & \ccAttribution{} & \ccAttribution{} & \ccAttribution{} \\
        GAIR & ViT-B/16 + MLP & Contrastive & Streetscapes1M & \ccAttribution{}\ccNonCommercial{} & \ccAttribution{} & \ccAttribution{}\ccShareAlike{} \\
        Climplicit & ResNet-18 + ReSIREN & Contrastive & CHELSA & \ccAttribution{} & \ccAttribution{} & \ccPublicDomain{} \\
        TIGeR & ViT-L/14 + MLP & Contrastive & \makecell{TIGeR-Train-4.5M \\ Cross-View Time \\ GLDv2, OSV-5M \\ MP-16} & \ccProprietary{} & \ccProprietary{} & \makecell{\ccProprietary{} \\ \ccUnknown{} \\ \ccAttribution{}\ccShareAlike{} \\ \ccAttribution{}\ccShareAlike{}\ccNoDerivatives{}\ccNonCommercial{}} \\
        LIANet & \makecell{ResNet-50 + U-Net + HT \\ ResNet-101 + U-Net + HT} & Generative & Sentinel-2 & \ccAttribution{} & \ccProprietary{} & \ccAttribution{} \\
        UniGeoCLIP & ViT-B/16 + Transformer & Contrastive & \makecell{Aerial imagery \\ street-level imagery \\ DSM, text, coords} & \ccAttribution{} & \ccProprietary{} & \ccProprietary{} \\
        TTE & ViT-L/16 + Spherical Voronoi & Contrastive & S2-100k & \ccAttribution{} & \ccAttribution{} & \ccAttribution{} \\
        \bottomrule
    \end{tabular}
    \label{tab:location-license}
\end{table}

Location encoders have the longest history, and many of the earliest models were released at a time when it was less routine to add a license to research products. The newest location encoders have the opposite problem, where the model code or weights are not yet publicly available. In particular, UniGeoCLIP was trained on a proprietary dataset, and thus the data and weights will never be released. While unknown or restrictive licenses make reproducibility challenging, unreleased and proprietary data and models make full reproducibility impossible. Table~\ref{tab:location-license} provides a full rundown of models, training datasets, and licenses for location encoders.

Understanding the full model provenance is not just an academic exercise; it can also have serious implications for the reliability and uncertainty of model predictions. Unlike explicit embeddings that are generally produced using models trained on globally distributed satellite imagery datasets, many location encoders are trained on datasets with known spatial bias. For example, models like GPS2Vec, GeoCLIP, and TIGeR are trained on datasets of geotagged images uploaded to sites like Flickr or Twitter, and are thus biased toward tourist-heavy regions. Models like CSP InceptionV3, SINR, and TaxaBind are trained on species occurrence records from iNaturalist, and are thus biased toward regions with better records, such as the United States and Europe. Models like GAIR, TIGeR, and UniGeoCLIP are trained on street-view imagery, and are thus biased toward road networks. LIANet is trained on satellite imagery, but only for Germany, and thus may not transfer to other regions.

\subsection{Patch embeddings}

\begin{table}[htbp]
    \centering
    \caption{\textbf{Reproducibility of patch embedding products available as of July 2026.} Data includes the data used during pretraining and the data used during inference to generate the embeddings. Licenses are broken down into the license for the model code, weights, data, and the resulting embeddings.}
    \begin{tabular}{@{}lccccccc@{}}
        \toprule
         & & & & \multicolumn{4}{c}{\textbf{Licenses}} \\
         \cmidrule(l){5-8}
        \textbf{Family} & \textbf{Model} & \textbf{Training} & \textbf{Data} & \textbf{Code} & \textbf{Weights} & \textbf{Data} & \textbf{Embed.} \\
        \midrule
         & & & Google Earth basemap \\
        \multirow{-2}{*}{MOSAIKS} & \multirow{-2}{*}{RCF} & \multirow{-2}{*}{Unsup.} & Planet basemap & \multirow{-2}{*}{\ccAttribution{}\ccNonCommercial{}} & \multirow{-2}{*}{\ccAttribution{}} & \multirow{-2}{*}{\ccUnknown{}} & \multirow{-2}{*}{\ccAttribution{}} \\
        \midrule
         & Clay v0 & & Landsat 8/9, NAIP, MODIS & & & \ccPublicDomain{} \\
         & Clay v1.5 & & NAIP & & & \ccAttribution{} \\
        \multirow{-3}{*}{Clay} & Clay v1.5 & \multirow{-3}{*}{MAE} & Sentinel-2 & \multirow{-3}{*}{\ccAttribution{}} & \multirow{-3}{*}{\ccAttribution{}} & \ccAttribution{} & \multirow{-3}{*}{\ccAttribution{}} \\
        \midrule
         & SoViT-400m/14 & SigLIP & WebLI, Sentinel-2 & \ccAttribution{} & \ccAttribution{} & \ccAttribution{}\ccShareAlike{} & \ccAttribution{} \\
         & ResNet-50 & DINO & Sentinel-2 & \ccAttribution{} & \ccAttribution{} & \ccAttribution{}\ccShareAlike{} & \ccAttribution{}\ccShareAlike{} \\
         & ResNet-50 & MoCo & Sentinel-1 & \ccAttribution{} & \ccAttribution{} & \ccAttribution{}\ccShareAlike{} & \ccAttribution{}\ccShareAlike{} \\
         & DINOv2-L/14 & DINOv2 & LVD-142M, Sentinel-2 & \ccAttribution{} & \ccAttribution{} & \ccAttribution{}\ccShareAlike{} & \ccAttribution{}\ccShareAlike{} \\
         & SoViT-400m/14 & SigLIP & WebLI, Sentinel-2 & \ccAttribution{} & \ccAttribution{} & \ccAttribution{}\ccShareAlike{} & \ccAttribution{}\ccShareAlike{} \\
         & ResNet-50 & DeCUR & Sentinel 1/2 & \ccAttribution{} & \ccAttribution{} & \ccAttribution{}\ccShareAlike{} & \ccAttribution{}\ccShareAlike{} \\
         & ConvNeXt V2 & MAE & MMEarth & \ccAttribution{} & \ccAttribution{}\ccNonCommercial{} & \ccAttribution{}\ccShareAlike{} \\
         & & & FLAIR-Hub, PASTIS-HD & & & \ccAttribution{} \\
         & & & TSAI-TS, Planted & & & \ccAttribution{} \\
         & & & S2-NAIP-Urban & & & \ccAttribution{} \\
         & & & EarthView-NEON & & & \ccAttribution{} \\
         & \multirow{-5}{*}{UniverSat} & \multirow{-5}{*}{MIM} & HyperGlobal & \multirow{-5}{*}{\ccAttribution{}} & \multirow{-5}{*}{\ccAttribution{}} & \ccAttribution{}\ccNonCommercial{}\ccShareAlike{} & \multirow{-5}{*}{\ccAttribution{}\ccShareAlike{}} \\
         & Clay v1.5 & MAE & see above & \ccAttribution{} & \ccAttribution{} & \ccAttribution{}\ccShareAlike{} & \ccAttribution{}\ccShareAlike{} \\
         & & & Sentinel 1/2 & & & \ccAttribution{} \\
         & & & Landsat 8, SRTM, CDL & & & \ccPublicDomain{} \\
         & & & WorldCereal & & & \ccAttribution{}\ccShareAlike{}\ccNonCommercial{} \\
         & & & WorldCover, Canopy Height & & & \ccAttribution{} \\
         & \multirow{-5}{*}{OlmoEarth-B} & \multirow{-5}{*}{MIM} & OSM & \multirow{-5}{*}{\ccAttribution{}\ccNonCommercial{}} & \multirow{-5}{*}{\ccAttribution{}\ccNonCommercial{}} & \ccAttribution{}\ccShareAlike{} & \multirow{-5}{*}{\ccAttribution{}\ccShareAlike{}} \\
         & ResNet-50 & SatCLIP & Sentinel-2 & \ccAttribution{} & \ccAttribution{} & \ccAttribution{}\ccShareAlike{} & \ccAttribution{}\ccShareAlike{} \\
         & ViT-B/16 & FarSLIP & RS5M, MGRS-200k, Sentinel-2 & \ccAttribution{} & \ccAttribution{} & \ccAttribution{}\ccShareAlike{} & \ccAttribution{}\ccShareAlike{} \\
        \multirow{-21}{*}{Major TOM} & SoViT-400m/14 & SigLIP & WebLI, Sentinel-2 & \ccAttribution{} & \ccAttribution{} & \ccAttribution{}\ccShareAlike{} & \ccAttribution{}\ccShareAlike{} \\
        \midrule
        Earth Index & ViT-S/16 & SoftCon & Sentinel-2 & \ccAttribution{} & \ccAttribution{} & \ccAttribution{} & \ccAttribution{} \\
        \midrule
        Copernicus & Copernicus-FM & MAE & Copernicus-Pretrain & \ccAttribution{} & \ccAttribution{} & \ccAttribution{} & \ccAttribution{} \\
        \bottomrule
    \end{tabular}
    \label{tab:patch-license}
\end{table}

Patch embeddings have the best reproducibility, with no proprietary models or data being used. Other than MOSAIKS, MMEarth, and OlmoEarth, which use custom licenses that prevent certain commercial applications, all model code and weights are distributed under permissive licenses that only require attribution. The Clay, Earth Index, and Copernicus embeddings all demonstrate excellent reproducibility, with the entire pipeline from pretraining and inference data to model code, weights, and resulting embeddings released under permissive licenses. Table~\ref{tab:patch-license} illustrates the full provenance from data to models to embeddings, alongside license information.

The largest set of patch embeddings, generated using existing model architectures and inference on Major TOM data, demonstrates an interesting reproducibility challenge. All Major TOM data is released under a copyleft CC-BY-SA-4.0 license, meaning that the resulting embeddings must also be released under the same share-alike license. This can make experimentation challenging for many companies, which tend to shy away from copyleft licenses because any derived products would also need to be released under the same license.

While all patch embeddings were generated using satellite imagery, many were generated using RGB-only subsets instead of complete multispectral imagery. The MOSAIKS embeddings were created using Google Earth and Planet RGB basemaps, while the Clay USA embeddings were created using RGB-only NAIP imagery. Similarly, the DINOv2, FarSLIP, and SigLIP Major TOM embeddings were created using only the RGB bands from Sentinel-2. Embeddings created using Sentinel-1 SAR imagery are likely better at certain applications like flood mapping. Only Copernicus-Embed was created using Sentinel 1/2/3/5P and Copernicus DEM, making it uniquely suited for oceanic and atmospheric applications like weather forecasting.

\subsection{Pixel embeddings}

\begin{table}[htbp]
    \centering
    \caption{\textbf{Reproducibility of pixel embedding products available as of July 2026.} Data includes the data used during pretraining and the data used during inference to generate the embeddings. Licenses are broken down into the license for the model code, weights, data, and the resulting embeddings.}
    \begin{tabular}{lccccccc}
        \toprule
         & & & & \multicolumn{4}{c}{\textbf{Licenses}} \\
         \cmidrule(l){5-8}
        \textbf{Product} & \textbf{Model} & \textbf{Training} & \textbf{Data} & \textbf{Code} & \textbf{Weights} & \textbf{Data} & \textbf{Embed.} \\
        \midrule
        Presto & Presto & MAE & \makecell{Sentinel 1/2 \\ ERA5, SRTM \\ Dynamic World} & \ccAttribution{} & \ccAttribution{} & \ccAttribution{} & \ccAttribution{} \\
        \midrule
        Tessera & Tessera & Barlow Twins & Sentinel 1/2 & \ccAttribution{} & \ccPublicDomain{} & \ccAttribution{} & \ccPublicDomain{} \\
        \midrule
         & & & Sentinel 1/2 & & & \ccAttribution{} \\
         & & & Landsat 8/9 & & & \ccPublicDomain{} \\
         & & & PALSAR ScanSAR & & & \ccAttribution{} \\
         & & & Copernicus DEM & & & \ccAttribution{} \\
         & & & GEDI & & & \ccPublicDomain{} \\ 
         & & & ERA5-Land & & & \ccAttribution{} \\ 
         & & & GRACE & & & \ccPublicDomain{} \\
         & & & NLCD & & & \ccPublicDomain{} \\
         & & & Wikipedia & & & \ccAttribution{}\ccShareAlike{} \\
        \multirow{-10}{*}{GSE} & \multirow{-10}{*}{AEF} & \multirow{-10}{*}{Distillation} & GBIF & \multirow{-10}{*}{\ccProprietary{}} & \multirow{-10}{*}{\ccProprietary{}} & \ccAttribution{} & \multirow{-10}{*}{\ccAttribution{}} \\
        \midrule
         & & & Landsat 5/7/8/9 & & & \ccPublicDomain{} \\
         & & & MODIS Terra & & & \ccPublicDomain{} \\
         & & & NASADEM & & & \ccPublicDomain{} \\
         & & & GAIA & & & \ccAttribution{}\ccNonCommercial{} \\
         & & & ESA WorldCover & & & \ccAttribution{} \\
        \multirow{-6}{*}{ESD} & \multirow{-6}{*}{ESDNet} & \multirow{-6}{*}{MTL} & GLAD & \multirow{-6}{*}{\ccProprietary{}} & \multirow{-6}{*}{\ccProprietary{}} & \ccAttribution{} & \multirow{-6}{*}{\ccAttribution{}} \\
        \bottomrule
    \end{tabular}
    \label{tab:pixel-license}
\end{table}

Unlike patch embeddings, which typically reuse foundation models produced by other authors, pixel embeddings are often jointly developed alongside new model architectures. This poses a new reproducibility challenge. Many of the models used to create pixel embeddings, including Google's AlphaEarth Foundations (AEF) and Embedded Seamless Data's ESDNet, are fully proprietary. This not only prevents scientists from reproducing these works, but also prevents them from building on them and making improvements. In comparison, Presto is released under a permissive license, and Tessera's model weights and embeddings are both public domain. Table~\ref{tab:pixel-license} lists the model architectures, pretraining techniques, datasets, and licenses for all known pixel embedding products.

The other challenge to reproducibility is data availability. No pixel embedding researchers have released stable checksummed downloads for the data used to pretrain their models or for embedding inference. Instead, they opt to release scripts that allow users to recreate the same data collection process, or simply release a list of source locations. However, due to frequent product updates with improved processing levels and cloud masking, this prevents perfect bit-for-bit reproducibility of the pretraining and inference data.

Similar to other embeddings, the data sources used during pretraining and inference also impact embedding quality and usefulness. Presto embeddings are limited to a small African nation. While Tessera has global 10~m resolution coverage for 2024, coverage for other years is only available on request. Google's satellite embeddings offer dense 10~m resolution coverage for all years for which the Sentinel missions have been active. While Sentinel and Landsat are used as model input, all other data sources are only used as ground truth, and thus the model may not adequately capture future changes to certain data sources. Embedded Seamless Data (ESD) sacrifices spatial resolution (30~m) for significantly longer temporal coverage (25 years) thanks to the long history of Landsat and MODIS missions. NASA data is typically public domain, ESA data is typically permissively licensed, and all other products depend on licenses for Wikipedia or GBIF contributions.

\section{Discussion}

Earth embeddings have brought a new wave of excitement and possibility to a number of important applications for environmental monitoring and understanding. However, they are still quite new, and practitioners will likely have a number of questions about embeddings. In this section, we delve into best practices for embeddings and end with open questions for future researchers to answer.

\subsection{Using existing embeddings}

Geospatial embeddings are ready for practical use. However, because the field is still evolving, we recommend standard remote-sensing validation and evaluation workflows. Treat them as reusable covariates, indexes, or initialization points whose value still has to be tested on the target task. A useful workflow starts by understanding your task's requirements for temporal cadence, spatial extent, and resolution: use location encoders when coarse geographic context is the signal, patch embeddings when the unit is a scene or search tile, and pixel embeddings when the label attaches to fields, parcels, or dense maps. Next, match temporal frequency to the phenomenon you are identifying. Snapshot products can work for land-cover or retrieval tasks, while annual and multi-year products are more defensible for crops, change detection, restoration, or other time-sensitive applications. Every analysis should compare embeddings against simple baselines and hand-designed covariates under spatial and temporal splits. Several studies show gains from embeddings, but the same evidence shows sensitivity to pooling, fusion, and transfer setting \citep{corley2026pixels,van2026better,ma2026harvesting}.

Current best practices are to be conservative: train simple downstream models first, inspect failure modes by region and class, and treat random train--test splits as insufficient evidence of transfer. For many remote-sensing tasks, zonal statistics are common; pixel embeddings follow the same pattern when users aggregate raster features over existing vector geometries. The choice of aggregation matters, and mean pooling can discard within-patch heterogeneity, while richer pooling can reduce geographic generalization gaps \citep{corley2026pixels}. For implicit and patch embeddings, check retrieval results against domain knowledge because neighborhoods in embedding space can be locally meaningful without supporting global vector arithmetic \citep{rahman2026characterizing}. Tools such as TorchGeo and rs-embed lower the engineering burden by exposing embedding products through common dataset interfaces \citep{torchgeo,ye2026any}. They are most useful when they preserve metadata downstream users need: coordinate reference systems, product versions, data provenance, temporal windows, sensor inputs, scaling transforms, quantization schemes, and missing-data conventions. As the number of embedding products continues to grow, agentic AI could prove beneficial to help users select one or more embeddings ideal for their application \citep{talemi2026agentic,munir2026agentic}.

\subsection{Publishing new embedding products}

For new embedding products, the central best practice is to make the product inspectable and reproducible at multiple levels. Authors should release model code, weights, training recipes, inference code, product metadata, and stable checksummed data references whenever licenses allow. If raw data cannot be redistributed, publish the exact collection versions, query parameters, cloud masks, dates, and spatial sampling frames. The embedding product should include a human-readable model card including the model name and version, sensor inputs, temporal window, coordinate reference system, grid definition, spatial resolution, data type, normalization or quantization transform, and license and intended usage.

The right storage format depends on a product's spatial structure: dense pixel embeddings suit cloud-optimized raster or array formats, while point, patch, and region embeddings suit vector or table formats (see Table~\ref{tab:formats}). The goal is routine reuse following FAIR principles: users can subset by space and time, read small windows without downloading full products, and recover meaningful numerical values from compressed storage.

\begin{table}[htbp]
    \centering
    \caption{\textbf{Recommended storage formats by embedding type.} A product's spatial structure decides the container. Point, patch, and region embeddings store one vector per feature and serialize to GeoParquet \citep{geoparquet}. Dense pixel embeddings suit cloud-optimized raster or array formats that support partial, windowed reads; GeoZarr \citep{geozarrspec} is preferable for annual or multi-year time series because GeoTIFF \citep{geotiff} has no native time dimension support without an accompanying sidecar file, such as a tile index. Location encoders tend to be compressed model representations of the Earth and can be served directly as model checkpoint files, such as ONNX.}
    \label{tab:formats}
    \begin{tabular}{llll}
        \toprule
        \textbf{Embedding type} & \textbf{Spatial structure} & \textbf{Temporal extent} & \textbf{Recommended format} \\
        \midrule
        Location / implicit & Irregular points & -- & ONNX or PyTorch \\
        Patch / region & Coarse tiles or polygons & -- & GeoParquet \\
        Pixel (dense) & Dense per-pixel grid & Snapshot & GeoTIFF or GeoZarr \\
        Pixel (dense) & Dense per-pixel grid & Time series & GeoZarr \\
        \bottomrule
    \end{tabular}
\end{table}

The type of embedding matters significantly when considering overall cost for creating an embedding product. For one year of coverage of Africa, roughly 30 million km\textsuperscript{2}, stored volume spans five orders of magnitude: from about 150~MB for the coarse Copernicus-Embed patch product to 77~TB for dense 10~m Presto pixel embeddings (see Table~\ref{tab:storage}) \citep{corley2026debt}. Patch embeddings are inexpensive to store and move but sacrifice spatial detail, while pixel embeddings preserve that detail at storage and egress costs that grow quickly, separating monthly storage bills of under a dollar from hundreds of dollars, and a single full-product download of a few dollars from thousands \citep{corley2026debt}.

\begin{table}[htbp]
    \centering
    \caption{\textbf{Storage cost of patch versus pixel embeddings.} Stored volume for one year of embedding coverage of Africa (\(\approx\)30 million km\textsuperscript{2}), using each product's published grid, dimensionality, and stored data type \citep{corley2026debt}. Estimated AWS costs represent storing one year of coverage for 12 months in S3 Standard in US West (Oregon), excluding request and data-transfer charges. AWS charges \$0.023/GB-month for the first 50 TB and \$0.022/GB-month for the next 450 TB. Patch products remain in the MB--GB range, while dense 10~m pixel products reach tens of TB and cost thousands of dollars per year to store.}
    \label{tab:storage}
    \begin{tabular}{lcrrcrr}
        \toprule
        \textbf{Product} & \textbf{Type} & \textbf{Resolution} & \textbf{Dimensions} & \textbf{Dtype} & \textbf{Storage} & \textbf{AWS S3/year} \\
        \midrule
        Copernicus-Embed & Patch & 0.25\textdegree{}\negphantom{\textdegree}\phantom{ km} & 768 & float32 & 147.5~MB & \$0.04 \\
        Clay v1.5 Sentinel-2 & Patch & 2.56~km & 1024 & float32 & 18.8~GB & \$4.83 \\
        Clay v1.5 NAIP & Patch & 256~m\phantom{k} & 1024 & float32 & 1.9~TB & \$488 \\
        Google Satellite Embedding & Pixel & 10~m\phantom{k} & 64 & int8 & 19.2~TB & \$4,935 \\
        Tessera & Pixel & 10~m\phantom{k} & 128 & int8 & 38.4~TB & \$9,871 \\
        Presto & Pixel & 10~m\phantom{k} & 128 & uint16 & 76.8~TB & \$19,497 \\
        \bottomrule
    \end{tabular}
\end{table}

Quantization and dimensionality reduction are important design decisions for reducing storage costs. Quantizing embeddings to lower precision, e.g., from float32 to int8, has been shown to cause negligible performance loss \citep{alphaearth,tessera}. Furthermore, using PCA to reduce to 64 dimensions combined with int8 yields 64\(\times\) compression with less than 2\% accuracy loss \citep{corley2026terrabit}. Binary quantization compresses the raw embedding payload a further 32\(\times\) and still recovers about 65\% of the true float32 nearest neighbors at planetary scale, enough for interactive retrieval and candidate generation \citep{corley2026terrabit}. Further investigation is necessary, however, because it is not well studied whether this quantization has a negative impact on fine-grained remote-sensing tasks.

\subsection{Coverage and uncertainty}

Current embeddings leave major parts of Earth observation underrepresented. Most public products are built from Sentinel, Landsat, or RGB imagery and target terrestrial land applications. This selection ignores hyperspectral imagery, thermal infrared, lidar, altimetry, ocean color, atmospheric composition, and other derived physical variables important for many applications. Copernicus-Embed is an important exception because it includes a broader set of Copernicus inputs, but its coarse resolution and weather-oriented design make it a different product class \citep{copernicus}. However, the lack of coverage over open ocean makes it less useful for global weather modeling and restricts its use to regional-level fine-tuning only.

The field needs embeddings that represent oceans, atmosphere, snow and ice, inland water, and human infrastructure with the same care now given to land-cover mapping. It also needs products that encode uncertainty. A single deterministic vector hides cloud contamination, seasonal aliasing, sensor gaps, interpolation, and model disagreement. Downstream users need quality masks, uncertainty layers, ensembles, or calibration diagnostics to know when an embedding should not be trusted \citep{zhu2024foundations,sialelli2026from}. This is especially important for proprietary models, where interpretability and explainability become even more challenging without access to model code or weights. 

\subsection{Benchmarking priorities}

Better benchmarks are the primary scientific bottleneck. An audit of geospatial foundation model papers found identical models reported with accuracy differences of more than ten points on the same benchmark, and most papers using training configurations that share no overlap with other work \citep{corley2026noone}. This implies that the field needs a shared, independently maintained evaluation harness and leaderboard rather than more one-off benchmarks. TorchGeo-Bench is an example of a toolkit that has recently been proposed as a solution to this problem \citep{torchgeobench2026}. 

The current landscape has many models and products, but they are often evaluated under different tasks, labels, spatial splits, and temporal windows. Future benchmarks should compare implicit, patch, and pixel embeddings under shared protocols for supervised mapping, regression, retrieval, spatial transfer, and temporal change. They should report simple baselines, label scale, train--test geography, temporal alignment, storage cost, inference cost, and licensing constraints alongside accuracy. Retrieval benchmarks should measure semantic correctness rather than only nearest-neighbor distance, and time-series benchmarks should evaluate the direction and timing of change rather than static classification alone. Such benchmarks would help distinguish which embedding products are robust across tasks, regions, and time from those that work only under narrow evaluation settings, while giving new products a clearer path for measurable improvement.

\paragraph{Acknowledgments}

The authors would like to thank Mikolaj Czerkawski (Major TOM, EarthEmbeddingExplorer) for reviewing a draft of this book chapter.

OpenAI Codex, powered by GPT-5.5 \citep{openai2026gpt55systemcard}, was used as an AI writing and research assistance tool during the preparation of the Introduction, Current Usage, and Experiments sections. Specifically, it was used to discuss chapter structure, improve readability, refine literature-synthesis text, and suggest candidate references related to geospatial embeddings. All AI-assisted text was reviewed and substantially revised by the authors; every retained sentence was checked by the authors, and all literature claims and citations were independently verified against the cited papers. The authors take full responsibility for the content, interpretation, and conclusions of the manuscript. Claude Opus 4.8 \citep{anthropic2026claudeopus48} was used for proofreading and revising the Discussion section with the intent to make it more readable and digestible to a wider audience. Claude Fable 5 \citep{fable5} was used to generate the TikZ-based code for Figure~\ref{fig:embedding-types}.

% \pagebreak

\printbibliography

\end{document}